\documentclass[10pt,twocolumn,letterpaper]{article}

\usepackage[pagenumbers]{wacv}      

\usepackage{graphicx}
\usepackage{amsmath}
\usepackage{amssymb}
\usepackage{booktabs}
\usepackage[accsupp]{axessibility}
\usepackage{setspace}
\graphicspath{ {images/} }

\usepackage[pagebackref,breaklinks,colorlinks]{hyperref}

\usepackage[capitalize]{cleveref}
\crefname{section}{Sec.}{Secs.}
\Crefname{section}{Section}{Sections}
\Crefname{table}{Table}{Tables}
\crefname{table}{Tab.}{Tabs.}

\def\wacvPaperID{1997} 
\def\confName{WACV}
\def\confYear{2024}

\begin{document}

\title{Automated Camera Calibration via Homography Estimation with GNNs}

\author{Giacomo D'Amicantonio,  Egor Bondarev,  Peter H.N. De With\\
Eindhoven University of Technology\\
{\{ \tt{\small{g.d.amicantonio, e.bondarau, p.h.n.de.with}} \} \tt\small @tue.nl}
}
\maketitle

\begin{abstract}
Over the past few decades, a significant rise of camera-based applications for traffic monitoring has occurred. Governments and local administrations are increasingly relying on the data collected from these cameras to enhance road safety and optimize traffic conditions. However, for effective data utilization, it is imperative to ensure accurate and automated calibration of the involved cameras. This paper proposes a novel approach to address this challenge by leveraging the topological structure of intersections.

We propose a framework involving the generation of a set of synthetic intersection viewpoint images from a bird's-eye-view image, framed as a graph of virtual cameras to model these images. Using the capabilities of Graph Neural Networks, we effectively learn the relationships within this graph, thereby facilitating the estimation of a homography matrix. This estimation leverages the neighbourhood representation for any real-world camera and is enhanced by exploiting multiple images instead of a single match. In turn, the homography matrix allows the retrieval of extrinsic calibration parameters. As a result, the proposed framework demonstrates superior performance on both synthetic datasets and real-world cameras, setting a new state-of-the-art benchmark.
\end{abstract}

\section{Introduction}
\label{sec:intro}
\begin{figure}
\centering
    \subfloat[Intersection 1]{\includegraphics[width=0.235\textwidth]{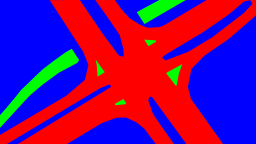}}\hfill
    \subfloat[Intersection 2]{\includegraphics[width=0.235\textwidth]{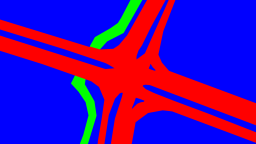}}\hfill
    \subfloat[Intersection 3]{\includegraphics[width=0.235\textwidth]{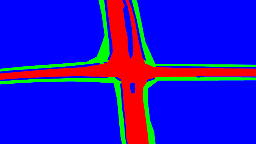}\label{fig:demo}}\hfill
    \subfloat[Intersection 4]{\includegraphics[width=0.235\textwidth]{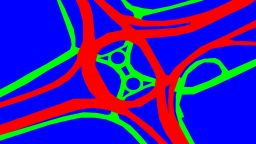}}\hfill
    \subfloat[Intersection 5]{\includegraphics[width=0.235\textwidth]{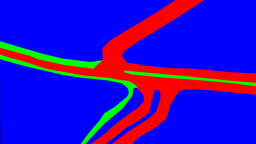}}\hfill
    \subfloat[Soccer field]{\includegraphics[width=0.235\textwidth, height=66pt]{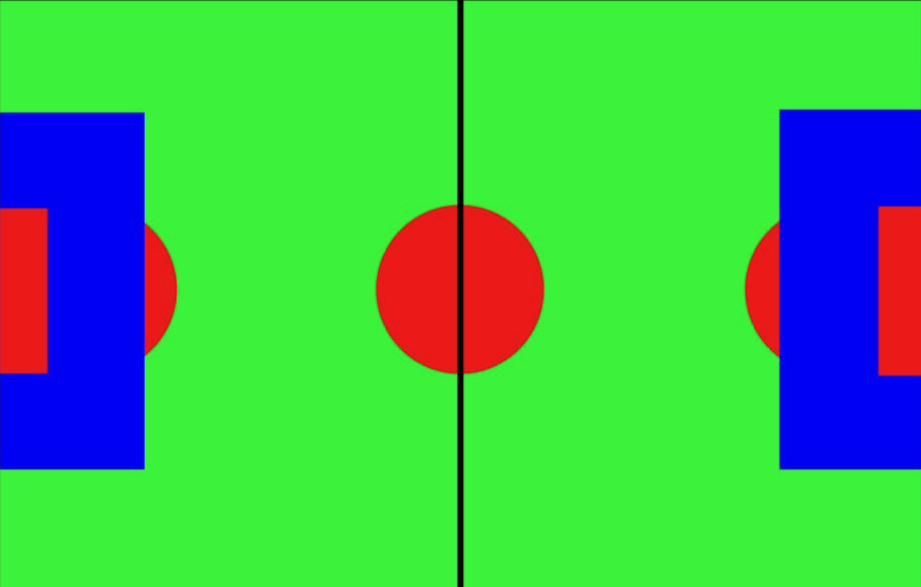}}\hfill
\caption{Intersection images from \cite{mine}. Bird's-eye-views of the semantical segmentations of the intersections: Red = road, Blue = terrain and Green = bicycle path. The intersections differ in the complexity of their topology. Semantic segmentation of the soccer field: Red = goal box, penalty arc and center circle, Blue = penalty box, Green = rest of the pitch. } 
\label{fig:bev}
\end{figure}
Camera calibration is a crucial aspect of computer vision (CV) applications, enabling the mapping of pixels to real-world coordinates. It serves as a prerequisite for various CV tasks, including object localization and immersive imaging. While calibrating cameras with a checkerboard pattern is now considered straightforward, this method is not always practical in real-world scenarios, especially when dealing with traffic-control cameras which are placed in busy intersections or in highways. These cameras are typically mounted on light posts and traffic lights, making them susceptible to slight movements caused by environmental factors and movement of large vehicles. Consequently, the original camera calibration may be affected by these movements, necessitating frequent and automated re-calibration. However, accessing these cameras in person is often unfeasible because of their locations at intersections or highways. 
\indent
While deep learning techniques have been explored to address this problem, they often demand substantial data and computational resources, posing limitations for startups and small companies. As of today, there is no industry standard for achieving cost-effective, accurate, and reliable automated camera calibration. To this end, this paper introduces a homography estimation approach that can be easily trained on synthetic data and performs effectively in real-world settings. \\
\indent
 The proposed method leverages the topological structure of intersections by creating a graph of synthetic templates from virtual cameras. Each template is generated by warping the bird's-eye-view (BEV) of the scene with a homography derived from each virtual camera. Using this graph, we find the best match between the camera image and a set of synthetic templates. This matching process is framed as a link-prediction task, executed by a Graph Neural Network (GNN) \cite{GNN1, GNN}. \\
 \indent
The proposed model predicts a probability score to select the top-k closest matches for the input image. The embeddings learned by the GNN are then processed to regress a homography that transforms the input image to the highest scoring template, whose homography serves as the anchor. The homography estimation is performed by a Spatial Transformer Network (STN) \cite{STN}.\\
\indent
Notably, this framework has two advantages. First, it estimates the homography between an image and the BEV by leveraging multiple views of the scene, rather than relying on a single image. Second, the framework is more computationally efficient than previous approaches, while outperforming them on synthetic datasets and real-world cameras. To the best of our knowledge, this is the first work to explore the use of GNNs for homography estimation.\\
\indent
The paper structure is as follows. Section 2 provides an overview of current camera calibration approaches and the literature on Graph Neural Networks. Section 3 details the models and the designed pipeline. The results on multiple intersections and five real-world cameras are presented in Section 4. Finally, Section 5 concludes the paper and indicates potential future work.
\begin{figure}[t]
\begin{center}
\includegraphics[width=0.45\textwidth]{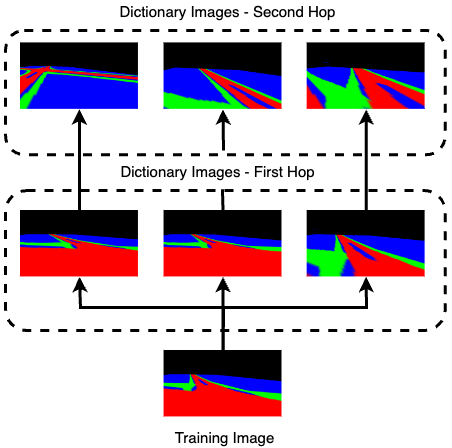}
\end{center}
   \caption{Example of the synthetic training and dictionary images. It is relevant to notice that in the second hop there can be images that are not visually similar to the training images or the representation of the corresponding part of the scene.}
\label{fig:neighours}
\end{figure}
\begin{figure}[t]
\begin{center}
\includegraphics[width=0.45\textwidth]{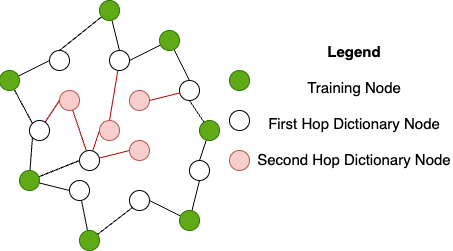}
\end{center}
   \caption{Reduced representation of a sub-graph of the generated graph. Each node represents an image and is connected to the 20 dictionary nodes with the lowest Topological Loss score.}
\label{fig:graph}
\end{figure}

\begin{figure*}[ht]
\begin{center}
\includegraphics[width=1\textwidth]{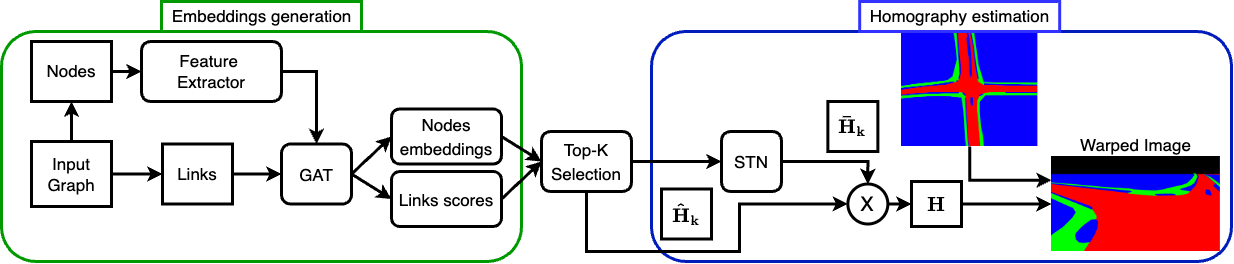}
\end{center}
   \caption{Overview of the proposed framework. In the first part, the features obtained from the nodes (images) in the sampled sub-graph are fed to the GAT network with the adjecency matrix of the links. The GAT produces embeddings for each node and predicts a score for each link. The top-k scoring links for each training (or testing) nodes are then processed by the STN to estimate the homography $\bar{\mathbf{H}}$.}
\label{fig:overview}
\end{figure*}

\section{Related Work}
\subsection{Camera Calibration}
Homography estimation typically involves exploitation of various image features and characteristics. The approaches mentioned in related work \cite{SIFT, SURF, GMS, LIFT, LPM, ORB, SOS} extract and match these features in images and then estimate the homography matrix via Direct Linear Transformation (DLT) \cite{DLT}. To handle erroneous matches, outlier detectors like RANSAC \cite{RANSAC} or the more recent MAGSAC, \cite{MAGSAC} are employed. However, these methods are not tailored specifically for traffic scenes and lack full automation.\\
\indent
In recent years, deep learning-based approaches have emerged as a powerful alternative. DeTone \etal \cite{detone} were among the first to propose a simple end-to-end network for direct homography estimation. Subsequent works have built further upon this foundation, using deeper and more complex networks to achieve better performance \cite{nowruzi, LE, content}. The introduction of the Spatial Transformer Network (STN) \cite{STN} sparked the development of new approaches, such as the works in \cite{UDH, SHA, mine}.
Addressing automated camera calibration, Bhardwaj \etal \cite{autocalib} presented a prominent method which exploits the presence of vehicles on the road, by employing vehicle-detection techniques, key-point extractors, and prior knowledge of the vehicle's geometrical properties. Another significant work in this area is reported in \cite{vehicleloc}, where the authors propose two camera calibration methods. One method optimizes an energy function to minimize the 3D re-projection error, while the other leverages synthetic data.
It is worth noting that although the mentioned approaches have shown success in various settings, none of them are specifically designed to handle the challenges unique to traffic scenes, \ie. the amount of vehicles in the scene and varying weather conditions.

\subsection{Graph Neural Networks} 
Since their introduction in \cite{GNN1, GNN}, Graph Neural Networks (GNNs) have demonstrated promising results in various tasks, involving graph-structured data. For instance, in \cite{classification}, Graph Convolutional Networks (GCNs) are employed for graph-level and node-level classification. Kipf and Welling \cite{VGAE} proposed an auto-encoder version of GNN for unsupervised link prediction. Over the years, GNNs have been employed for tasks such as nearest-neighbour estimation \cite{GMC} and sentiment analysis \cite{sentiment}.
Recent years have shown a growing interest in exploring the application of GNNs in image and video-related tasks. The graph embeddings learned by GNNs have facilitated various downstream tasks, including pose estimation \cite{e2efm, graphcovis} and object detection \cite{odam}. In \cite{keypoints}, the key points of an image are treated as nodes in a graph, with links connecting them, and a GNN model is trained to match these key points. Similarly, in \cite{superglue}, graphs representing key points are used for image matching. 
The successes of GNNs in image-related tasks have inspired us to apply this approach to the specific problem of automated camera calibration in traffic scenes. By constructing a graph of synthetic templates based on the topological structure of intersections, we train a GNN model to perform homography estimation. The use of GNNs sets our work apart from previous approaches and allows accurate automated camera calibration in traffic monitoring scenarios.

\begin{figure*}[ht]
\begin{center}
\includegraphics[width=1\textwidth]{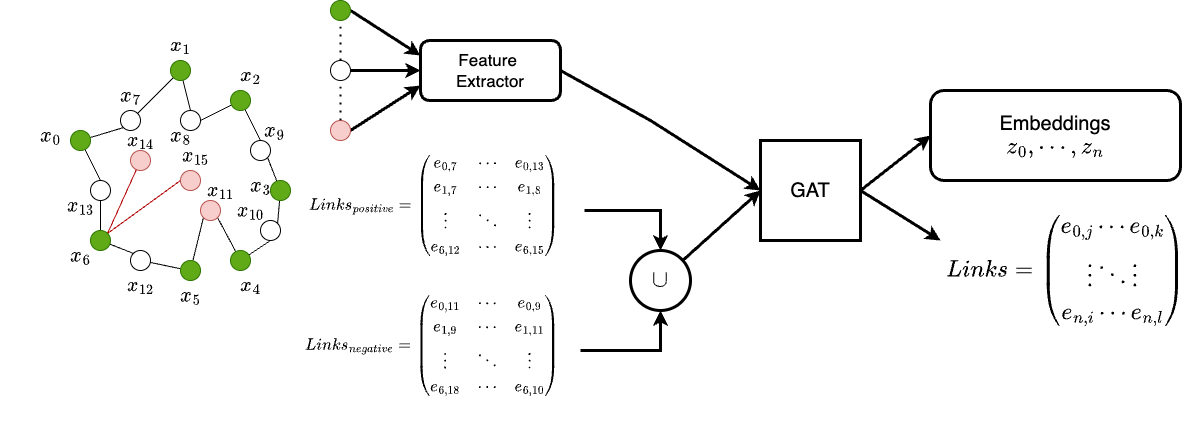}
\end{center}
   \caption{Overview of the embedding generation process. The features of all nodes in the mini-batch are generated by the feature extractor. The positive links are the links between each training node and its dictionary nodes in the mini-batch, while the negative links are all the links between training nodes and the other dictionary nodes. Merging positive and negative links creates the matrix of all possible links in the mini-batch. The GAT produces embeddings for each node and probability scores for the links, the latter being used to train both models via Equation~(\ref{eqn:bce}).}
\label{fig:gnn+fe}
\end{figure*}

\section{Method}
The proposed framework leverages a GNN to match the input image with a dictionary of templates, assigning a score to all possible links. A Spatial Transformer Network (STN) estimates the homography transformation between the highest scoring template and the input image from the embeddings generated by the GNN. The images and the templates are generated, starting from the manually segmented bird's-eye-view (BEV) of the intersection. Whereas we train the framework only on synthetic images, its performance is evaluated on a test set of synthetic images and on five real-world cameras placed in Intersection 3 (see Figure~\ref{fig:demo}).

\subsection{Data Generation}
Following the approach described in \cite{SHA} and \cite{mine}, we generate virtual cameras by randomly sampling intrinsic and extrinsic parameters from a grid. The semantically segmented BEV of the scene, shown in Figure~\ref{fig:bev}, is warped with the homography matrices of these virtual cameras, resulting in a set of approximately 20,000 images per scene.
To create the training and testing splits, images are randomly sampled from the generated set, with an equal distribution among the two splits. Additionally, a dictionary split is sampled containing images to be leveraged as anchors for the final homography estimation task, following \cite{SHA}.
Each split forms a graph, where each image corresponds to a node connected to the top-k most similar images in the dictionary (empirically, k=20). The similarity score between images can be calculated using similarity metrics, such as Topological Loss \cite{mine}. 
During both training and testing, nodes from the respective graph are sampled to construct mini-batches. As illustrated in Figure~\ref{fig:graph}, for each node in a mini-batch, we also sample a number of its neighbours at a 2-hop distance, following the approach presented in~\cite{hamilton2017inductive}. However, we discard all nodes in the second hop that are not part of the dictionary to prevent any cross-contamination between splits. As shown in Figure~\ref{fig:neighours}, the images in the second hop can be visually different from the training image, or representing a different part of the intersection.  \\

\subsection{Matching Process}
In our approach, a light-weight feature extractor derives a feature vector from each image. 
In addition to the feature vectors, a matrix is constructed containing all possible links between nodes in the mini-batch. Both the feature vectors and the matrix of links serve as inputs to a Graph Attention Network (GAT) to perform the matching task. The GAT model has been proposed in~\cite{gatv2} and represents an improvement over the preceding design reported in~\cite{gat}. GATs follow the same principles as GCNs, but with a key difference in the way they handle neighbourhood features. While a standard GCN normalizes node features across neighbours during the convolutional operation, a GAT employs a masked attention mechanism to compute attention coefficients between nodes within the same neighbourhood, thereby preserving important structural information.\\
\indent
The feature extractor and the GAT are trained jointly, allowing the former to learn meaningful representations of the original images for the latter. The training loss function used for the GAT is the conventional Binary Cross-Entropy, defined as:
\begin{equation}
L_{GAT} = -\sum_{n=0}^{N-1}{\big(\,y_{n}\log(p_{n}) + (1 - y_{n})\log(1 - p_{n})\,\big)}.
\label{eqn:bce}
\end{equation}
Here, $N$ represents the number of possible links, $y_{n}$ is the binary label for link $n$, and $p_{n}$ is the score predicted by the GAT for that link. This loss function ensures effective training of both the feature extractor and the GAT for accurate and robust link prediction.\\
\indent
An overview of the link-prediction task is shown in Figure~\ref{fig:gnn+fe}. The final outputs of the GAT are the embeddings for every node in the batch and a score for each possible link between these nodes.

\begin{figure*}[ht]
\begin{center}
\includegraphics[width=1\textwidth]{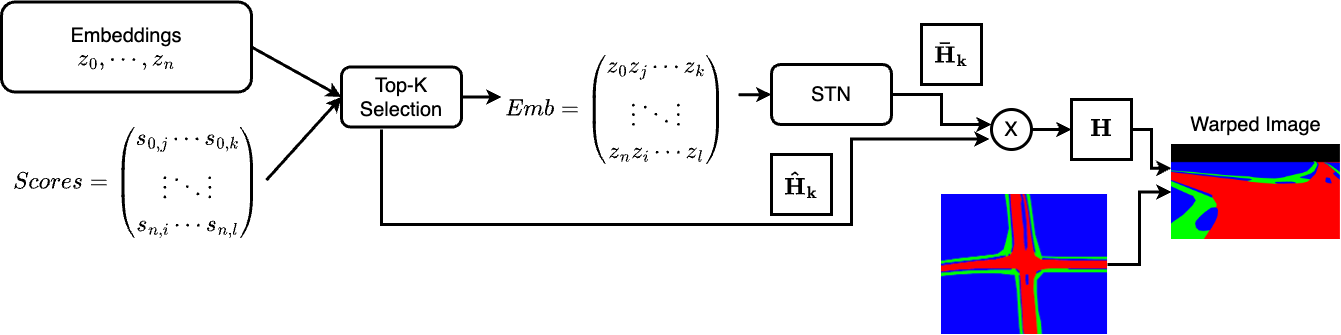}
\end{center}
   \caption{Overview of the homography estimation task. For each training node, the probability scores predicted by the GAT are needed to select the top-k best nodes, whose embeddings are then used by the STN to predict $\bar{\mathbf{H}}$.  
   The homography of the node connected to a training node via the highest scoring link is the anchor for the final estimated homography that warps the BEV. The resulting warped images are employed to train end-to-end all models via Equation \ref{eqn:tl}.}
\label{fig:gnn+stn}
\end{figure*}

\subsection{Homography Estimation}
The GAT predicts a score for every possible link between training and dictionary nodes in the batch. The top-k scoring links for each training (or testing) node are utilized to select the neighbours containing the most useful information for the Spatial Transformer Network (STN). The feature vectors are concatenated into a two-dimensional feature matrix and passed as input to the homography estimation component. The homography of the highest-scoring dictionary node, $\hat{\mathbf{H}}_k$, is the anchor for the corresponding training (or testing) node. In fact, the STN estimates a homography that transforms the best match found by the GNN to the target image. In other words, the homography $\bar{\mathbf{H}}$, estimated by the STN, is multiplied with its anchor $\mathbf{\hat{H}_k}$, producing the final homography $\mathbf{H} = \mathbf{\hat{H}_k}\bar{\mathbf{H}}$.\\
\indent
The STN described in~\cite{mine} is adapted to accept the embeddings produced by the GAT. Following that work, we adopt the same training strategy: the semantic BEV is warped by the estimated homography and the resulting warped image is compared to the synthetic ground truth via a Topological Loss implementation of the MSE. This is described by Equation~(\ref{eqn:tl}), where the running indexes $k, l\in\{-1, 0, +1\}$ and $i, j\in\{0, ..., \sqrt{N}\}$, while $Y$ and $\hat{Y}$ are the ground-truth image and the warped image, respectively.
\begin{align}
\begin{split}
\label{eqn:TLij}
    \mathcal{L}_{\text{patch}}(\hat{Y}, Y) &= \text{MSE}(\hat{Y}_{i,j}, Y_{i,j}) + \\ \nonumber
    & \alpha\sum_{k}\sum_{l}\max{(0, \text{MSE}(\hat{Y}_{i+k,j+l}, Y_{i+k,j+l})-\beta}),
\end{split}
\end{align}
\begin{equation}
\label{eqn:tl}
    \text{and}\ \ \mathcal{L}_{\text{Top-MSE}} = \frac{1}{N}\sum_{i}\sum_{j}\mathcal{L}_{\text{patch}}(\hat{Y}_{i,j}, Y_{i,j}).
\end{equation}

\begin{table*}[ht]
\caption{Obtained IoU scores by different implementations of the proposed model, using three different GNNs and the baseline models, over the five synthetic intersection datasets and the World Cup 2014 dataset. All values (mean + std) are expressed in \%.}
\centering
\begin{tabular}{l|ccccc|c}
    \toprule
    \textbf{Model} & \multicolumn{5}{c|}{\textbf{Intersections}} & \textbf{World Cup}\\
     & \textbf{1} & \textbf{2} & \textbf{3} & \textbf{4} & \textbf{5} & \textbf{2014} \\
    \midrule
     Sha \etal \cite{SHA} & 85.96 & 78.42 & 78.99 & 83.16 & 71.95 & 81.64\\
     D'Amicantonio \etal \cite{mine} & 87.91 & 87.00 & 86.66& 84.51 & 73.75 & 84.21 \\
     \midrule
     Proposed 1 (GCN \cite{classification}) & 92.31$\pm$0.37 & 93.25$\pm$0.16 & 93.58$\pm$0.11 & 88.28$\pm$0.36 & 87.99$\pm$0.21 &  89.91$\pm$0.44\\
     Proposed 2 (GAT \cite{gat}) & 95.61$\pm$0.99 & \textbf{96.57}$\pm$0.68 & 93.98$\pm$1.45 & 88.89$\pm$1.2 & 96.27$\pm$0.24 & 96.89$\pm$0.7 \\
     Proposed 3 (GATv2 \cite{gatv2}) & \textbf{95.96}$\pm$0.84 & 96.56$\pm$0.53 & \textbf{95.38}$\pm$1.32 & \textbf{89.59}$\pm$0.14 & \textbf{96.86}$\pm$0.57 & \textbf{97.31}$\pm$0.39 \\
     \bottomrule
\end{tabular}
\label{tab:testing}
\end{table*} 

\begin{figure*}
\centering
\begin{subfigure}[b]{0.41\textwidth} 
    \includegraphics[width=\linewidth]{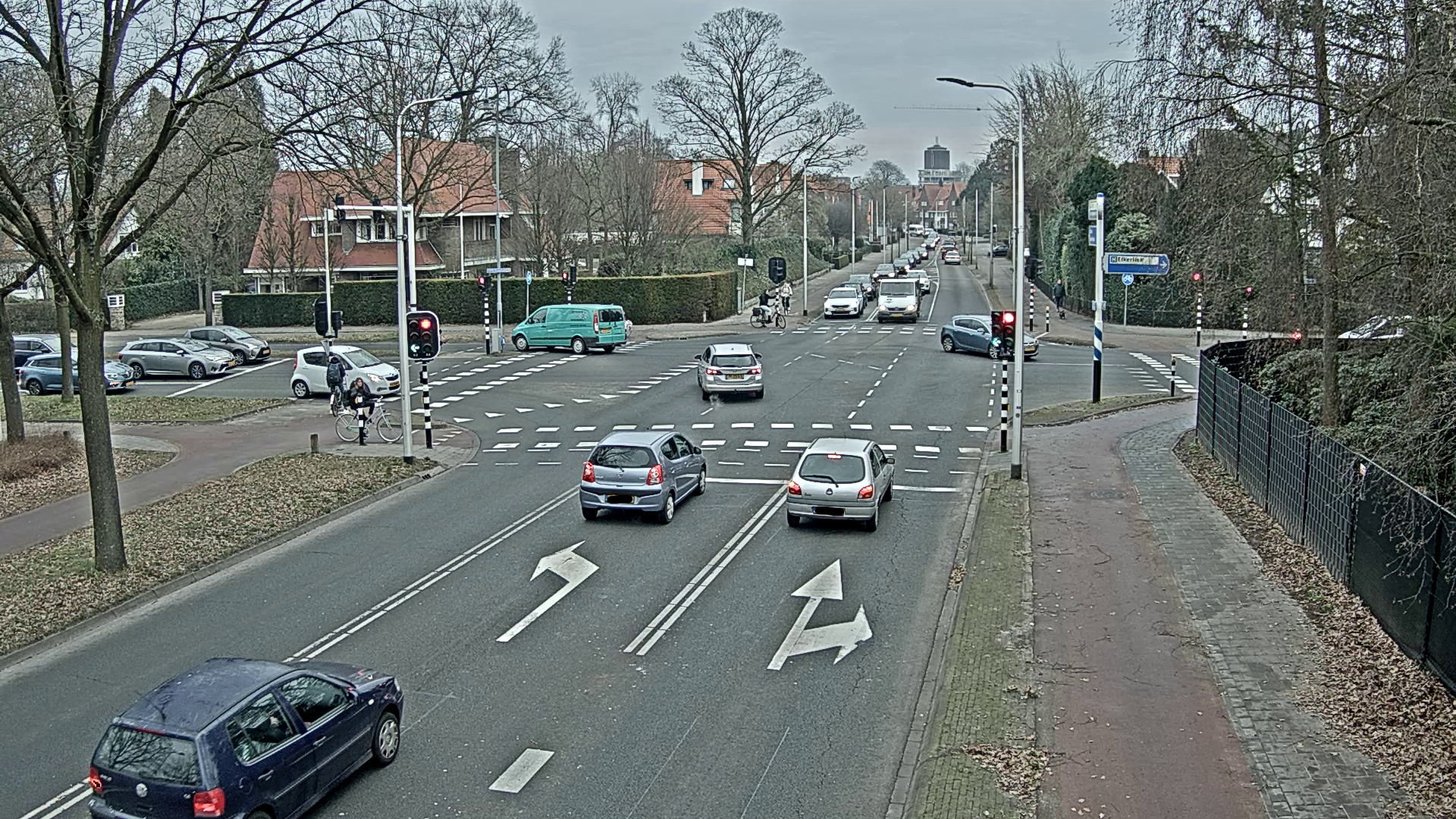}
    \caption{Original image.} 
    \label{fig:input}
\end{subfigure}%
\quad
\begin{subfigure}[b]{0.41\textwidth} 
    \centering
    \includegraphics[width=\linewidth]{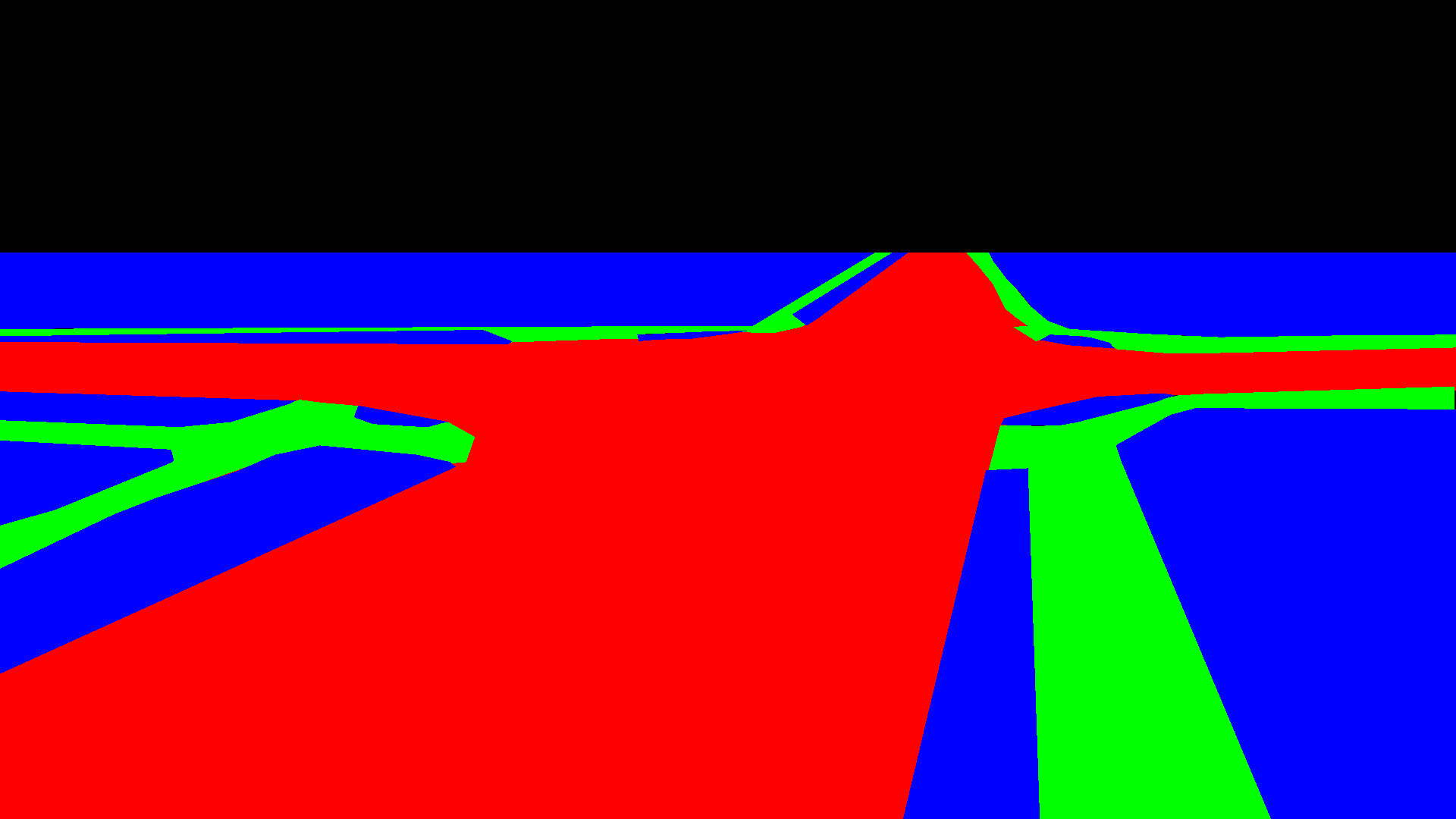}
    \caption{Ground truth image.} 
    \label{fig:gt}
\end{subfigure}
\begin{subfigure}[b]{0.41\textwidth} 
    \includegraphics[width=\linewidth]{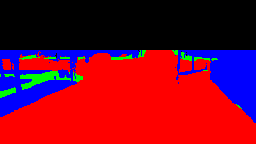}
    \caption{Segmented image.} 
    \label{fig:segm}
\end{subfigure}%
\quad
\begin{subfigure}[b]{0.41\textwidth} 
    \includegraphics[width=\linewidth]{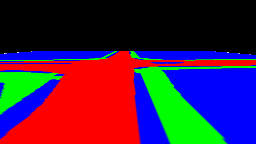}
    \caption{Warped image generated by the proposed model.} 
    \label{fig:warp}
\end{subfigure}%
\caption{Visual results of segmentation and warping steps for a real-world image. The segmented image obtained from the SegFormer is less precise than the synthetic images on which the model was trained, which partially explains the drop in accuracy between testing images and real-world images.} 
\label{evalimg}
\end{figure*}

\begin{table*}[ht]
\caption{Comparison between the proposed model with different feature extractors. All values (mean + std) are expressed in \%.}
\centering
\begin{tabular}{l|ccccc|c}
    \toprule
    \textbf{Model} & \multicolumn{5}{c|}{\textbf{Intersections}} & \textbf{World Cup}\\
     & \textbf{1} & \textbf{2} & \textbf{3} & \textbf{4} & \textbf{5} & \textbf{2014}\\
    \midrule
     Proposed (GATv2 + ResNet18) & 95.13$\pm$0.13 & 95.75$\pm$0.37 & 94.45$\pm$0.17 & 87.65$\pm$0.61 & 95.97$\pm$0.07 & 95.41$\pm$0.44 \\
     Proposed (GATv2 + 4-Conv) & \textbf{95.96}$\pm$0.84 & \textbf{96.56}$\pm$0.53 & \textbf{95.38}$\pm$1.32 & \textbf{89.59}$\pm$0.14 & \textbf{96.86}$\pm$0.57 & \textbf{97.31}$\pm$0.39 \\
    \bottomrule
\end{tabular}
\label{tab:compres}
\end{table*} 

\begin{table}[t]
\caption{Average IoU scores obtained by the proposed models and the baseline models on five real-world cameras located in Intersection 3. All values (mean + std) are expressed in \%.}
\centering
\begin{tabular}{l|c}
    \toprule
    \textbf{Model} & \textbf{Cameras} \\
     \midrule
     Sha \etal \cite{SHA} & 76.44$\pm$3.56 \\
     D'Amicantonio \etal \cite{mine} & 81.12$\pm$2.54 \\
     \midrule
     Proposed 1 (GCN \cite{classification}) & 85.73$\pm$0.34 \\
     Proposed 2 (GAT \cite{gat}) & 86.48$\pm$0.85 \\
     Proposed 3 (GATv2 \cite{gatv2}) & \textbf{87.89}$\pm$0.93 \\
    \bottomrule
\end{tabular}
\label{tab:evaluation}
\end{table} 

\section{Results}
\subsection{Experimental setup}
The GAT and its feature extractor are trained jointly until convergence, using the loss for link prediction reported in Equation~(\ref{eqn:bce}). In this way, the GAT learns the features of the nodes and their neighbours which can be better exploited by the STN. After this warmup period of 30~epochs, we start training the full pipeline of feature extractor, GAT and STN, using only the loss function in Equation~(\ref{eqn:tl}). Therefore, the GAT may learn to choose nodes outside the neighbourhood of the training node as more relevant for the training objective. The IoU scores are measured between the ground-truth images and the BEV warped with the estimated homographies to evaluate the model. If the IoU score does not improve for 10~epochs, the learning rates of all models are halved, and if it does not improve for a further 15~epochs the training is stopped. \\
\indent
We evaluate the model on five different intersection scenes and the World Cup 2014 dataset~\cite{worldcup}. The results shown in Table~\ref{tab:testing} are averaged over five training cycles for each dataset and different train/test/dictionary splits are sampled at every cycle. Table~\ref{tab:evaluation} validates the results by evaluating the model performance on images obtained from five cameras located in Intersection 3. The experimental results are compared to the baselines, reported in \cite{mine} and \cite{SHA}. Furthermore, the performances are compared with the original GAT model proposed by~\cite{gat} and with a standard GCN presented in~\cite{classification}. 

\subsection{Ablation study}
Table~\ref{tab:testing} shows that the proposed method outperforms the state-of-the-art models with any of the three GNN networks, strongly supporting the idea of a graph-based approach for this task. The main difference between the proposed work and the competitors is in the way the structural information of the intersection is leveraged. In fact, both previous works employed a Siamese model to perform the matching task, for which the best template from the dictionary was selected. This approach has two downsides, as listed below. 
\begin{itemize}
  \item Selecting a single match is valid but not optimal: the amount and quality of topological information that becomes available from a single match is inferior to the information that can be extracted from multiple matches and the relationships between them.
  \item The Siamese network as a matching component is computationally expensive: each input image is compared to every image in the dictionary, which places a heavy burden on computational resources and hinders the scaling capabilities of the model.
\end{itemize}
Framing the matching process as a link-prediction task in a graph solves both problems, allowing us to retrieve multiple matches at a lower computational cost.
Furthermore, the GNN learns features for the nodes based on the related visual features and the relationships between them, which gives the STN a much richer input. Training the models end-to-end has been proven not effective in~\cite{mine}, due to the design of the matching process. In this work, training all models end-to-end provides much more flexibility to the GNN, allowing it to discard subpar matches directly connected to the training node in favor of other nodes in the neighbourhood that can be more relevant for the STN. \\
\indent
It should be noticed that the difference between the three types of GNN we experimented with, do not show a significant difference in performance. The GCN proves to be the least performing of the three, but it still outperforms the current state-of-the-art by 10\% on average. Moreover, the difference in performance between GAT and GATv2 is minor.\\
\indent
The employed feature extractor to prepare the images for the GNN models is a lightweight model, comprising four convolutional layers only. To prove that such a simple model is not a bottleneck for our framework, Table~\ref{tab:compres} compares the performances with a larger and more complex feature extractor such as ResNet18. The larger feature extractor does not have a positive impact on the overall performance of the model, as it is outperformed by the much simpler model on all datasets.  We conjecture that because of the much simpler nature of the semantic images, a more complex model is harder to train due to the higher amount of parameters. 

\subsection{Evaluation on real camera views}
To validate the results obtained by the proposed model on synthetic images, the models are executed on images obtained from five cameras located in Intersection 3. The images are segmented by SegFormer~\cite{segformer}, which is trained on Cityscapes~\cite{cityscapes}. Given that Cityscapes has 30~classes while our application only needs 4~(terrain, road, bicycle path and background), we discard or merge the additional classes in one of the four classes. While the merging strategy heavily depends on the individual scene, it is fair to assume that e.g. the car class can be merged with the road class and the bicycle class with the bicycle-path class.\\
\indent
In Table~\ref{tab:evaluation}, the proposed model is compared with the previous baseline model. With each of the three GNNs, the proposed model outperforms the baseline models by up to 8\%. It is worth noting that the segmented images are not as precise as the synthetic images, which may cause the drop in accuracy w.r.t. the testing dataset. Furthermore, we do not account for any distortion while generating the synthetic images, which may be a factor for real-world cameras, especially fish-eye cameras. As shown in Figures~\ref{fig:segm} and~\ref{fig:gt}, the segmentation obtained from the SegFormer model is not very precise in some areas, such as the bicycle path at the right or the poles and trees at the left. We conjecture that fine-tuning a segmentation model to discard obstructing objects may lead to even better results with the same model. We leave this for future work, as it is not in the scope of this paper.
\section{Conclusions}
In this work, we have proposed a homography estimation framework based on Graph Neural Networks. It is shown that framing multiple synthetic views of an intersection as a graph and training a GNN to learn the relationships between them, leads to the generation of rich embeddings which can be useful to estimate the homography of a given image. The framework is completely automated and requires fewer computational resources and training data w.r.t. comparable methods in literature. The proposed framework has been tested on five different intersections, five real-world cameras and a real-world dataset with three different types of GNNs. In all settings, the framework outperforms the current state-of-the-art by a significant margin, proving to be also very effective in real-world applications. 
In future work, it can be interesting to focus on improving the scoring mechanism for the matches retrieved by the GNN and refining the segmentation of camera images.

{\small
\bibliographystyle{ieee_fullname}
\bibliography{egbib}
}

\end{document}